\documentclass[onecolumn]{cohere}

\usepackage{lipsum}

\usepackage{amsmath,amsfonts,bm}

\def\eqref#1{equation~\ref{#1}}

\def\1{\bm{1}}

\DeclareMathAlphabet{\mathsfit}{\encodingdefault}{\sfdefault}{m}{sl}
\SetMathAlphabet{\mathsfit}{bold}{\encodingdefault}{\sfdefault}{bx}{n}

\usepackage{stfloats} 
\usepackage{hyperref}
\usepackage{url}
\usepackage{booktabs}
\usepackage{graphicx}
\usepackage{multirow}
\usepackage[export]{adjustbox}
\usepackage{float}
\usepackage{caption}
\usepackage{subcaption}
\usepackage[colorinlistoftodos]{todonotes}
\usepackage{lscape}
\usepackage{rotating}
\usepackage{enumitem}
\usepackage{wrapfig}
\usepackage{tabularx}
\usepackage{chessfss}

\title{Elo Uncovered: \\ Robustness and Best Practices in \\ Language Model Evaluation}

\author{
    name={Meriem Boubdir},
    affiliation={Cohere for AI},
    email={meri.boubdir@gmail.com}
}
\author{
    name={Edward Kim},
    affiliation={Cohere},
    email={edward@cohere.com}
}
\author{
    name={Beyza Ermis},
    affiliation={Cohere for AI},
    email={beyza@cohere.com}
}
\author{
    name={Sara Hooker},
    affiliation={Cohere for AI},
    email={sarahooker@cohere.com}
}
\author{
    name={Marzieh Fadaee},
    affiliation={Cohere for AI},
    email={marzieh@cohere.com}
}

\date{\today}
\abstract{
In Natural Language Processing (NLP), the Elo rating system, originally designed for ranking players in dynamic games such as chess, is increasingly being used to evaluate Large Language Models (LLMs) through ``A vs B'' paired comparisons.
However, while popular, the system's suitability for assessing entities with constant skill levels, such as LLMs, remains relatively unexplored. 
We study two fundamental axioms that evaluation methods should adhere to: \emph{reliability and transitivity}. We conduct extensive evaluation of Elo behaviour, illustrating that individual Elo computations exhibit volatility and delving into the  impact of varying the Elo rating system's hyperparameters.
We show that these axioms are not always satisfied raising questions about the reliability of current comparative evaluations of LLMs.
If the current use of Elo scores is intended to substitute the costly head-to-head comparison of LLMs, it is crucial to ensure the ranking is as robust as possible.
Guided by the axioms, our findings offer concrete guidelines for enhancing the reliability of LLM evaluation methods, suggesting a need for reassessment of existing comparative approaches.
}

\begin{document}

\section{Introduction}

In the rapidly evolving field of Natural Language Processing (NLP), the task of accurately and reliably evaluating LLMs has become increasingly challenging~\citep{liang2022holistic,chang2023survey,srivastava2023beyond,kaddour2023challenges,pozzobon2023challenges}.
Human feedback has emerged as an indispensable tool in this performance assessment process, serving as a qualitative metric that captures nuances that automated scoring mechanisms often fail to address ~\citep{askell2021general,bai2022training,bai2022constitutional,srivastava2023beyond,ding2023enhancing,dettmers2023qlora}.

These human-centered evaluations, highly valuable to the overall progress of the NLP field, typically adopt an \textit{``A vs B''} comparative setup, turning evaluations into a zero-sum game between language models. This paired feedback structure~\citep{zhao2023slichf} naturally lends itself to the Elo rating system, originally designed for ranking chess players for better matchmaking~\citep{elo1978rating}.
With Elo rating system, we can integrate subjective human feedback into a structured rating system and assess the performance of language models.

\begin{figure*}[t!]
  \centering
    \includegraphics[width=1.\linewidth]{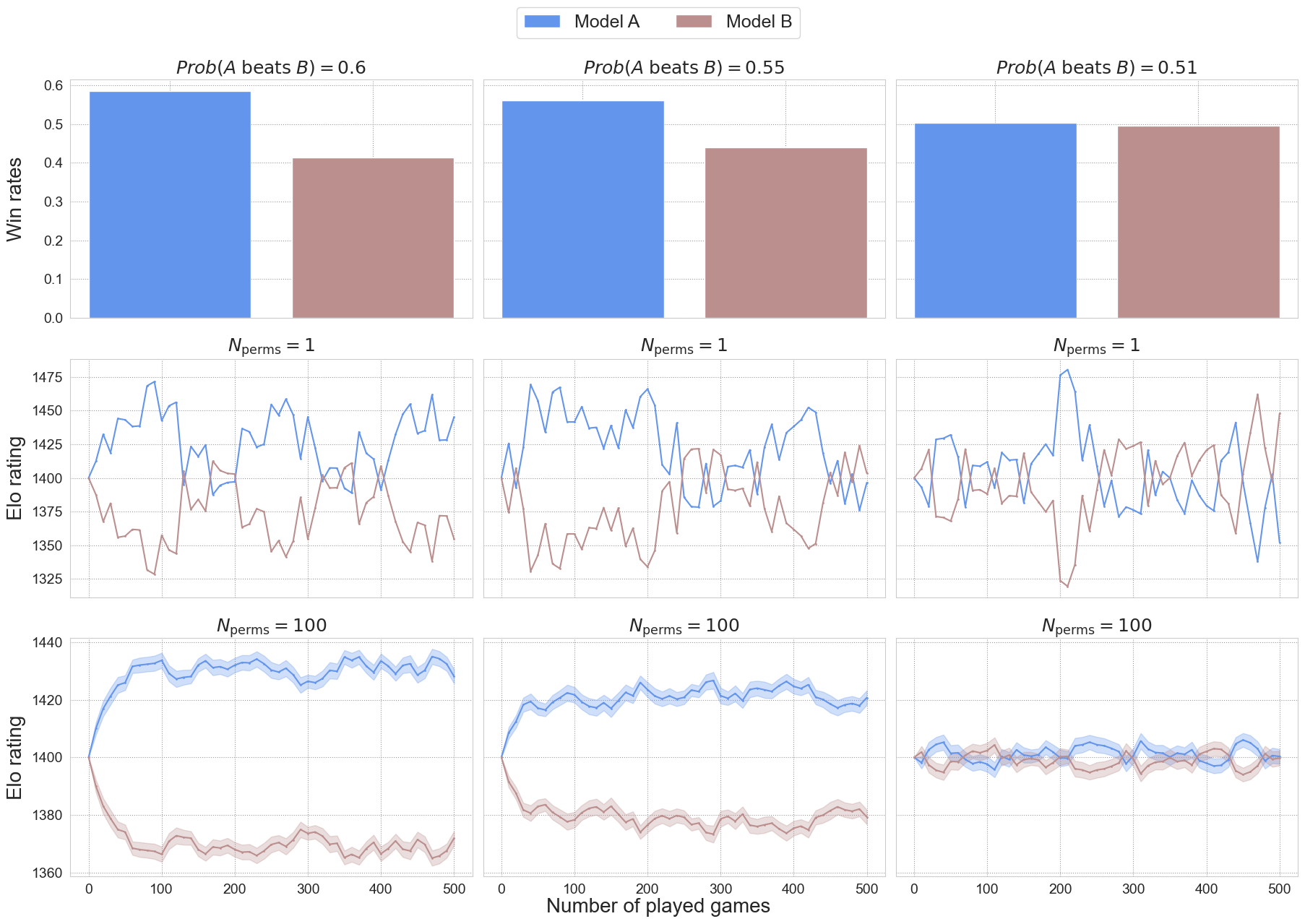}
\caption{\textbf{Impact of win probabilities and permutation sampling on Elo ratings}: Comparing Model A and Model B across three different win probabilities (\(Prob(A \; \mathrm{{beats}} \; B) = \{0.6, 0.55, 0.51\}\)) with two levels of permutation sampling (\(N_{\text{{perms}}}=1\) and \(N_{\text{{perms}}}=100\)). The top row displays the observed win rates, the middle row illustrates Elo ratings with a single permutation, and the bottom row shows the mean and standard error of the mean (SEM) of Elo ratings across 100 permutations.}
\label{fig:ordering-impact}
\end{figure*}

The core principles of Elo rating have proven to be resilient and adaptable due to its dynamic adjustments, relative rating focus, consistency across skill levels, and simplicity and transparency.
As a result, the Elo system has found diverse applications, from predicting sports events outcomes \citep{Binder2009TheEO,Hvattum10football,LEITNER2010471,Wise2021EloRF}, 
and facilitating matchmaking in massively multiplayer online games like StarCraft \textsc{ii} and Dota ~\citep{Ebtekar2021EloMMRAR,esportsheadlines,liquipedia,eslpro}, to its recent use in the evaluation of LLMs \citep{askell2021general,bai2022training,bai2022constitutional,srivastava2023beyond,ding2023enhancing,dettmers2023qlora,chatArena,lin2023llmeval}. However, to-date there has not been a comprehensive examination of the compatibility of Elo scores and LLMs.

Unlike dynamic competitors that evolve, LLMs have static capabilities and operate in a time-agnostic context. In this setting, not only are LLM evaluations unconstrained by a preset number of turns (unlike tournament timelines or predefined match sequences), but the ordering of matches can also significantly influence the final Elo scores and, consequently, models rankings.
This oversight is especially concerning, given the direct impact of Elo system rankings on both research directions and real-world applications in NLP as well as its widespread adoption \citep{ye2023large,zheng2023judging,koepf2023openassistant,askell2021general,bai2022training,bai2022constitutional,srivastava2023beyond,ding2023enhancing,dettmers2023qlora,chatArena,lin2023llmeval}.

This study aims to close this research gap by adopting an axiomatic approach and scrutinizing both the reliability and limitations of the Elo rating system when applied to LLMs. We study two fundamental axioms that evaluation methods should adhere to: \emph{reliability and transitivity}. Through theoretical and empirical analyses grounded in collected human feedback data, our contributions provide a comprehensive understanding of when and how to reliably employ the Elo system for LLM evaluation, thus offering valuable guidelines for researchers and practitioners in the NLP field.

We find that Elo ratings for LLMs are highly sensitive to the order of comparisons and the choice of hyperparameters. Moreover, desirable properties such as transitivity are not always guaranteed, and can be unreliable unless there is comprehensive human feedback data for all unique pairwise comparisons among models in the feedback pool. The sensitivity of Elo ratings becomes more pronounced when dealing with models that exhibit similar performance levels. 
We illustrate the best practices for addressing Elo rating sensitivities by offering guidelines for hyperparameter selection and matchmaking scenarios.

\textbf{Implications of our work} As LLMs rapidly advance, evaluation leaderboards are gaining popularity to assess the performance of newly introduced models using Elo scores. 
Elo can also be used in the learning framework of LLMs to produce a ranking of models and their outputs for preference training.
No research has explored the nuances of using Elo scores to compare LLMs, which, unlike chess, exhibit static capabilities and operate in a time-agnostic manner.
We show that Elo rating does not always satisfy two critical axioms---reliability and transitivity---leading to rankings of models that are not accurate. Our research offers guidelines for reliable and robust implementation of Elo scores when comparing LLMs. Deviation from our recommendations could result in inaccuracies when ranking LLMs, particularly in situations where model performances are closely matched, and Elo differences are minimal (a common occurrence in many real-world scenarios).

\section{Elo Algorithm Explained}
\label{sec:elo-system}

We provide the mathematical formulation of the Elo algorithm, contextualized to the setting of LLM evaluation. In this formulation, let \( \mathcal{M} \) be a set of models and each model \( i \in \mathcal{M} \) is assigned an initial numerical rating \( R_i \).

\subsection{Expected Score Computation}
For a given paired match-up between two models \( A \) and \( B \) (\(A, B \in \mathcal{M} \)), each with respective ratings \( R_A \) and \( R_B \), the expected scores \( E_A \) and \( E_B \) are computed as:
\begin{subequations}
    \begin{align}
    E_A &= \frac{1}{1 + 10^{(R_B - R_A) / 400}} \\
    E_B &= \frac{1}{1 + 10^{(R_A - R_B) / 400}}
    \end{align}
\end{subequations}

In this context, the factor of \(400\)~\citep{elo1978rating} precisely adjusts the sensitivity of the expected score to differences in ratings.
A \(400\)-point advantage in ratings translates to a \(10:1\) odds in favor of the higher-rated model, providing an interpretable metric for performance comparison.
For evenly matched models \((R_A = R_B)\), both \(E_A\) and \(E_B\) equate to \(0.5\), reflecting a \(50:50\) win probability for both models.

\subsection{Rating Update Mechanism}
Following each match, the Elo ratings are updated based on the observed outcome.
The rating adjustment is dictated by the equation:
\begin{equation}
    R'_A = R_A + K(S_A - E_A)
    \label{eq:elo-update}
\end{equation}
Here, \( S_A \) represents the actual score achieved by model \( A \), which can take on either the value 0 or 1.
The \( K \)-factor serves as a variable hyperparameter to adapt the rate of change in rating to different scenarios.

\section{Desirable Properties of Elo}

The objective of using Elo scores to rank models is to establish a comparative understanding of the performance hierarchy among them.
When incorporating a new model into an already ranked list, only a limited number of pairwise annotations are required to determine its position in the ranking. 
The ability to infer the relative performance of a model in comparison to all previous models in the list relies on the robustness of the scoring method and the transitive property of the ranking.
We describe this desirable characteristics for an evaluation framework using two axioms: \emph{transitivity and reliability}.

\subsection{Axiom 1: Transitivity}

A desirable property of any rating system is transitivity because it ensures consistency and logical coherence in the way entities are ranked or rated.
Transitivity in this context means that if player \( A \) beats player \( B \), and player \( B \) beats player \( C \), then player \( A \) is expected to beat player \( C \). 
If the ranking of large language models exhibits transitivity, we can deduce their comparative performance without the need for direct head-to-head evaluations between every pair of models.
The central assumption in the development of various leaderboards for comparing language models is that the rankings adhere to the principle of transitivity~\citep{zheng2023judging}.

While Elo's design inherently assumes transitivity, our synthetic data, which are derived from realistic scenarios, uncovers certain circumstances that violate this assumption.
Such anomalies can subsequently affect the final ranking of language models and their relative performance assessments.

\subsection{Axiom 2: Reliability} 

We consider two aspects of reliability:

\textbf{Sensitivity to ordering:} Unlike chess or time-bound sports where match sequences are structured, in LLM evaluations all matches can occur independently and in parallel, amplifying the sequence's influence on final models ranking.
If the prompts are presented in a specific order, and one model happens to perform better on the initial set of prompts, it may gain an advantage in subsequent comparisons due to the cumulative effect of its early success.
This inherent variability prompts us to investigate the extent to which match-up ordering affects the robustness of Elo ratings. 

\textbf{Sensitivity to hyperparameters:} The sensitivity of hyperparameters can compromise the robustness of elo scores leading to inconsistent rankings.
Evaluating and understanding this sensitivity is crucial for building evaluation frameworks that maintain consistency across diverse models.
In this work, we evaluate the sensitivity of Elo performance to one key hyperparameter, the \(K\) factor. This factor acts as a scaling constant in the Elo rating system, pivotal for updating ratings after each matching. It essentially determines how quickly a model's rating converges to what can be considered its ``true'' skill.
While conventional applications like chess use standard \(K\)-factor values (16 for experienced players and 32 for novices), these may not be directly applicable in the context of evaluating LLMs due to the unique characteristics and requirements of this domain.

\begin{figure}[t]
  \centering
    \includegraphics[width=.65\linewidth]{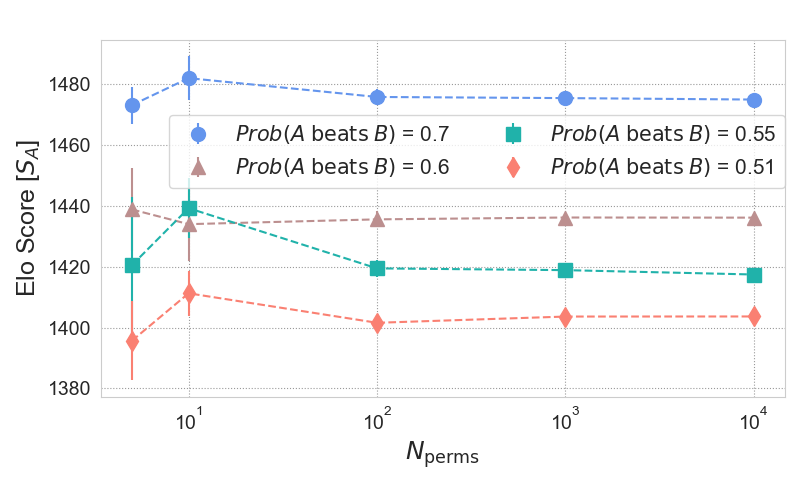}
  \caption{Variation of Model A's average Elo score with increasing number of permutations (\(N_{\text{perms}}\)) for different probabilities of Model A winning (\(Prob(A \; \mathrm{{beats}} \; B) \)). Error bars indicate standard errors of the mean.}
  \label{fig:nperms_var}
\end{figure}

\section{Synthetic Human Feedback}
\label{sec:synth-hf}

Given the costly and time-consuming nature of human evaluations, studying the Elo system's behavior under various scenarios becomes challenging. 
To circumvent these limitations, we first validate properties of Elo using synthetic data generation through Bernoulli processes to simulate various human feedback scenarios. In \ref{sec:appl} we will extend these evaluations to include real-world human feedback.
This time-agnostic and independent setup of LLM evaluations resembles a  Bernoulli process~\citep{bernoulli1713ars}, a sequence of independent experiments, each yielding a simple ``win'' or ``loss'' outcome, representing one model outperforming another. We use this setting where we can control characteristics of the distribution to evaluate different desirable properties of a rating system. 

In this controlled setting, our primary objectives include testing the \textbf{transitivity} axiom---whether a consistently higher-rated model outperforms those with lower ratings in all scenarios. Additionally, in studying the \textbf{reliability} axiom, we explore how the Elo scores are affected by the \textit{order in which models are compared} and the sensitivity to \textit{hyperparameter adjustments}, particularly the \(K\)-factor. This synthetic setup offers a robust platform to dissect and understand the dynamics of the Elo rating system in the context of LLM evaluations, without the constraints and limitations of relying solely on real-world human feedback.

\subsection{The Bernoulli Analogy}
Pairwise comparisons in LLM evaluation draw parallels with the foundational principles of the Bernoulli experiment in probability theory. 
This section delves into the similarity between human feedback-based evaluations and the Bernoulli experiment's principles.
\subsubsection{Preliminaries}
A Bernoulli trial is a random experiment with exactly two possible outcomes, ``success'' or ``failure''. These outcomes adhere to the condition:
\begin{equation}
P(A) + P(A^c) = 1
\end{equation}
Here, the random variable \(\mathcal{X}\) denotes the outcome, where \(\mathcal{X} = 1\) implies success, and \(\mathcal{X} = 0\) signifies failure. The probabilities are:
\begin{equation}
\begin{aligned}
    P(\mathcal{X} &= 1 ) &= p , \quad &
    P(\mathcal{X} &= 0 ) &= 1 - p 
\end{aligned}
\end{equation}
with \(0 \leq p \leq 1\), the ``success'' probability.

\subsubsection{Mapping to Human Feedback}
When comparing two models, \(A\) and \(B\), across \(N\) pairwise evaluations, the setup aligns with a Bernoulli process. 
This process comprises a sequence of independent and identically distributed (\textit{i.i.d}) Bernoulli trials.
To frame this analogy, we designate a win probability, \(P(A_{\text{win}})\), to model \(A\). Leveraging a Bernoulli random variable, \(\mathcal{X}\), as a means to simulate synthetic human feedback, we proceed as follows:
\begin{enumerate}[topsep=0pt, partopsep=0pt]
    \item A sample is drawn from \(\mathcal{X}\) using \(P(A_{\text{win}})\).
    \item If \(\mathcal{X} = 1\), feedback suggests a preference for model \( A \).
    \item Otherwise, model \(B\) is favored.
\end{enumerate}

\subsubsection{Extending to Multiple Players}
Given a finite set of \(n\) distinct models \(\mathcal{M}\), their pairwise comparisons can be formulated as:
\begin{equation}
\label{eq:combination}
\binom{n}{2} = \frac{n!}{2!(n-2)!}
\end{equation}
This formula yields \(\binom{n}{2}\) unique pairs \( (A,B) \) where \( A, B \in \mathcal{M} \) and \( A \neq B \). 
For each of these pairs, a Bernoulli process, comprising multiple Bernoulli experiments, is conducted to discern which model performs better over a sequence of trials.

\subsection{Synthetic Data Generation}
Building upon the Bernoulli process analogy, when conducting multiple independent evaluations between two models, the distribution of the number of times one model is preferred over the other naturally follows a binomial distribution. For \(N\) pairwise comparisons, the relation is:
\begin{equation}
P(k; N, p) = \binom{N}{k} p^k (1-p)^{N-k}
\end{equation}
where \(P(k; N, p)\) is the probability of one model being preferred \( k \)
times out of \( N \) evaluations.
\( p \) is the success probability and \( \binom{N}{k} \) is the binomial coefficient, representing the number of ways to choose \(k \) successes from \( N \) trials.


\begin{figure}[t]
  \begin{subfigure}{.48\textwidth}
    \centering
    \caption{Elo Scores for a Single Sequence}
    \includegraphics[width=.95\linewidth]{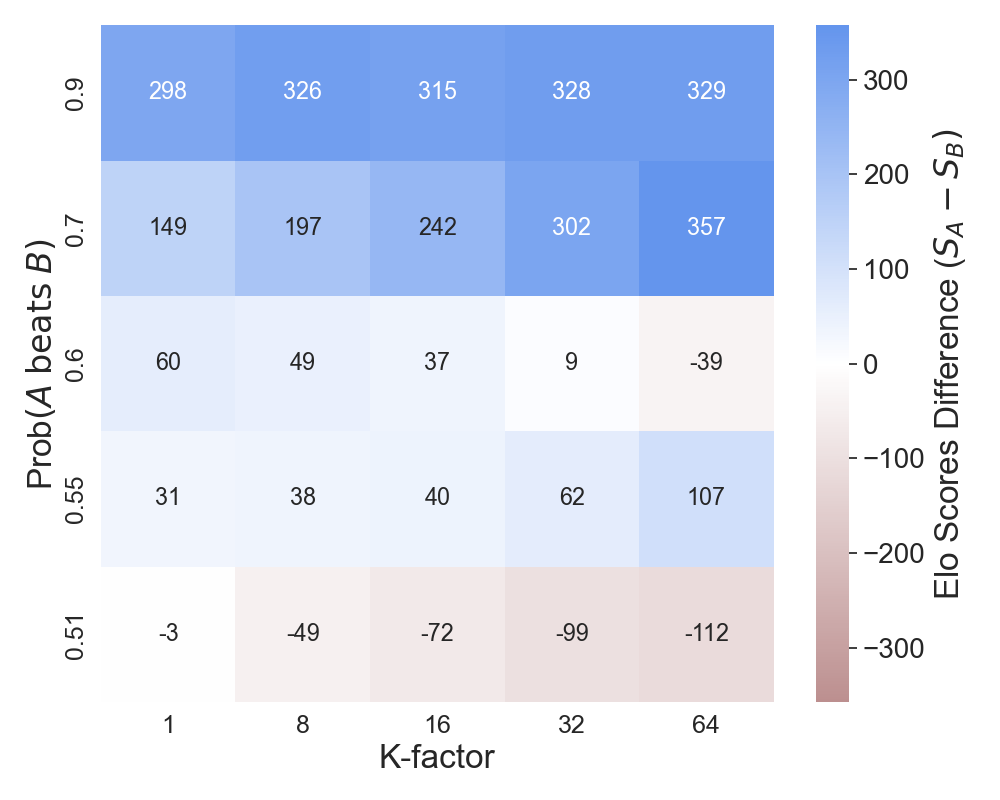}
  \end{subfigure}
  \hfill 
  \begin{subfigure}{.48\textwidth}
    \caption{Elo Scores Averaged Over 100 Permutations}
    \includegraphics[width=.95\linewidth]{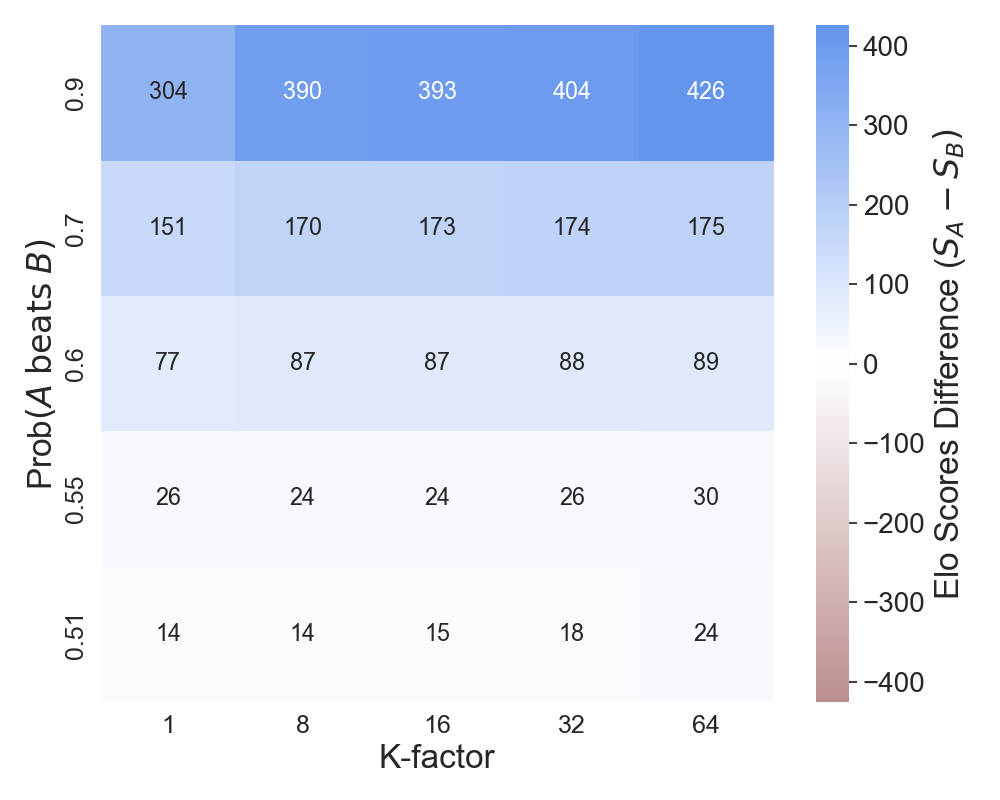}
  \end{subfigure}
\caption{Final Elo scores difference (\(S_A - S_B\)) as a function of \(K\)-factor and \(N_{\text{perms}} \). Positive values reflect the expected ranking where Model \( A \) is superior to Model \( B \), while negative values indicate a discrepancy, falsely suggesting that Model \( B \) has a higher Elo score than Model \( A \). We compare between a single sequence of outcomes and averages over \(N_{\text{{perms}}} = 100 \) unique permutations.}
  \label{fig:heatmaps}
\end{figure}

\section{How Robust Are Elo Scores?}
\label{sec:stress-tests}

This section defines rigorous stress tests designed to investigate the whether the two axioms are satisfied in this evaluation framework.
We focus on critical desirable properties of a ranking mechanism -- that it should 1) be insensitive to match-up ordering, 2) not be overly sensitive to hyperparameters like \(K\)-factor 3) preserve properties of transitivity. Subsequently, we provide empirically-grounded guidelines for safe and interpretable application of Elo ratings.

\subsection{Impact of Ordering on Elo Ratings}

\subsubsection{Experimental Setup} 
To quantify the effect of match-up ordering on Elo ratings, we generate a baseline sequence of \( N_{\text{games}} = 1000 \) match outcomes between models \( A \) and \( B \), reflecting the scale typical of LLM evaluations via human feedback.
We hold \( N_{\text{games}}\) constant for the entirety of
our study to maintain consistency.
From this baseline, we derive \( N_{\text{perms}}\) distinct permutations, each involving a complete reshuffling of the initial sequence to simulate various chronological orders in which the games might unfold.
It is important to note that we are not generating new match outcomes for each permutation; instead, we simply reorder the existing data to explore the potential impact of different match-up sequences.
For each reordered sequence, we update the Elo ratings \(R_A\) and \(R_B\) according to equation \ref{eq:elo-update}, resetting both ratings to an initial value of 1400 at the start of each permutation.
Following this, we compute average Elo ratings per match across all \( N_{\text{perms}} \) permutations, ensuring a robust analysis that takes into account the full range of possible match-up orders.

We compare ratings' behavior for a set of selected winning probabilities 
\(Prob(A \; \mathrm{{beats}} \; B) \) = \{0.51, 0.55, 0.6\}, inspecting a spectrum of real-world scenarios. 
\( N_{\text{perm}}\) is varied from a minimum of 1 to a maximum of 10k, providing a robust sample size for statistical analysis (see Figure~\ref{fig:nperms_var}).
Subsequently, we compute the average Elo ratings per match across all permutations.
These averages, \textbf{\( \Bar{R}_{A}\)} and \textbf{\( \Bar{R}_{B}\)}. particularly for \(N_{\text{perms}} = 1\) and  \(N_{\text{perm}} = 100\), are visualized to offer insights into the stability of the ratings, as shown in Figure~\ref{fig:ordering-impact}.

\subsubsection{Key Findings}
Our analysis underscores the interplay between winning probability \(P(A_{\text{win}})\) and the number of different orderings \(N_{\text{perm}}\) on the stability of Elo ratings after each update.
For \(P(A_{\text{win}}) \geq 0.6\), Elo ratings demonstrate high stability; additional results for \(P(A_{\text{win}}) = 0.65\) and beyond are available in Appendix~\ref{apdx:elo-ordering-impact}.
On the other hand, for \(P(A_{\text{win}}) \approx 0.5\), ratings exhibit significant instability for a single sequence.
As depicted in Figure~\ref{fig:ordering-impact}, when both models have a win probabilities are around \(0.5\), Elo ratings frequently intertwine, making it challenging to discern a clear performance difference between the two.
The instability plateaus as \(N_{\text{perms}}\) exceeds 100, resulting in stabilized Elo ratings that align closely with the preset winning probabilities.
For instance, at \(P(A_{\text{win}}) = 0.55\), the average Elo rating for Model \(A\), \(\Bar{R}_{A}\), consistently exceeds that for Model \(B\), \(\Bar{R}_{B}\), when averaged across multiple permutations, reflecting an accurate performance-based ranking of these models.

These observations validate our concerns highlighted earlier, emphasizing the critical role of \(N_{\text{perms}}\) for a  reliable interpretation of Elo ratings in LLM evaluations. In Elo-based evaluations, the sequence of which models are compared is not a mere procedural detail; it can significantly influence the final Elo scores. 
In scenarios involving models of similar quality and capabilities, which is often the case, this sensitivity is exacerbated.

\subsection{Sensitivity to Hyperparameters}

\subsubsection{Experimental Setup}
We extend our previous approach by conducting tests across a range of winning probabilities and multiple \(K\)-factor values (\(1, 8, 16, 32, 64\)).
We compute and compare the average Elo scores \(\Bar{S}_A\) and \(\Bar{S}_B\) for \(N_{\text{{games}}} = 1000\) and \(N_{\text{{perms}}} = \{1, 100\}\).
The differences between these final averages for Model \(A\) and Model \(B\) are summarized in Figure~\ref{fig:heatmaps} to assess the stability and expected ranking between the two models.

\vspace{-1.94em}

\subsubsection{Key Findings}
As shown in Figure~\ref{fig:heatmaps}, notable instability is observed in model rankings based on the final Elo scores when we consider a single sequence of paired comparisons (i.e., \(N_{\text{{perms}}} = 1\)), especially for winning probabilities nearing 0.5.
This instability is markedly exacerbated at higher \(K\)-factors. In contrast, the picture changes when coupling higher \(K\)-factors with raising the number of permutations to at least 100.

Higher \(K\)-factors, in this multi-permutation scenario, speed up the differentiation between models' Elo scores, enabling faster convergence to their true skill levels. This yields much more stable and reliable model rankings. It is noteworthy that this faster convergence is observed to be more reliable for higher winning probabilities, which corresponds to skewed win rates in a real-wold scenario.

\begin{table}[t]
\centering
\caption{Investigation of Elo score reliability in capturing true model hierarchies across varying configurations. Scenarios explore the transitive relationship \(A > B \; \text{and} \; B > C \implies A > C\). The star (\textcolor{red}{*}) indicates cases where the Elo score fails to accurately reflect the expected hierarchy of models. \(\approx\) represents models with similar performance; \(\gg\) indicates that a model significantly outperforms the other one.}
\begin{tabular}{|c|c|c|c|c|c|}
\hline
\multirow{2}{*}{Scenario} & \multirow{2}{*}{Model} & \multicolumn{4}{c|}{Models Ranking per Configuration} \\
\cline{3-6}
&& \(N=1, K=1\) & \(N=100, K=1\) & \(N=1, K=16\) & \(N=100, K=16\) \\
\hline
\multirow{3}{*}{
  \begin{tabular}{c}
    \textsymking \\ 
    \(A \gg B\) \\ \(B \gg C\)
  \end{tabular}
} 
& \(A\) & 1539.43  & 1528.50 \(\pm\) 0.35 & 1650.93  & 1584.78 \(\pm\) 3.09 \\
& \(B\) & 1390.47  & 1410.33 \(\pm\) 0.54 & 1381.17  & 1406.48 \(\pm\) 3.23 \\
& \(C\) & 1270.10  & 1261.17 \(\pm\) 0.33 & 1167.90  & 1208.74 \(\pm\) 2.71 \\
\hline
\multirow{3}{*}{
  \begin{tabular}{c}
    \textsymrook \\ 
    \(A \gg B\) \\ \(B \approx C\)
  \end{tabular}
} 
& \(A\) & 1502.09  & 1495.92 \(\pm\) 0.36 & 1509.08  & 1526.04 \(\pm\) 3.03 \\
& \(B\) & 1337.48  & \textcolor{red}{\textbf{1342.70*}} \(\pm\) 0.53 & 1379.00  & 1340.83 \(\pm\) 2.83 \\
& \(C\) & 1360.42  & \textcolor{red}{\textbf{1361.38*}} \(\pm\) 0.38 & 1311.92  & 1333.13 \(\pm\) 2.68 \\
\hline
\multirow{3}{*}{
  \begin{tabular}{c}
    \textsymbishop \\ 
    \(A \approx B\) \\ \(B \gg C\)
  \end{tabular}
} 
& \(A\) & 1437.97  & \textcolor{red}{\textbf{1433.84*}} \(\pm\) 0.41 & 1440.31  & 1460.22 \(\pm\) 2.90 \\
& \(B\) & 1455.10  & \textcolor{red}{\textbf{1453.84*}} \(\pm\) 0.61 & 1481.04  & 1452.87 \(\pm\) 3.25 \\
& \(C\) & 1306.93  & 1312.32 \(\pm\) 0.34 & 1278.65  & 1286.91 \(\pm\) 2.72 \\
\hline
\multirow{3}{*}{
  \begin{tabular}{c}
    \textsymknight \\ 
    \(A \approx B\) \\ \(B \approx C\)
  \end{tabular}
}
& \(A\) & 1426.33  & 1419.73 \(\pm\) 0.36 & 1407.44  & 1432.26 \(\pm\) 2.93 \\
& \(B\) & 1390.47  & 1393.29 \(\pm\) 0.59 & 1386.17  & 1392.75 \(\pm\) 3.04 \\
& \(C\) & 1383.20  & 1386.99 \(\pm\) 0.41 & 1406.39  & 1374.99 \(\pm\) 3.12 \\
\hline
\end{tabular}
\label{tab:elo-trans-synth}
\end{table}

\subsection{Transitive Properties of Elo Scores}

\subsubsection{Experimental Setup}
The transitivity property of the Elo scores is defined as:
\begin{equation}
    A > B \quad \text{and} \quad B > C \implies A > C
\end{equation}

To test the transitivity property, we design four distinct scenarios that model real-world conditions:

\begin{enumerate}[topsep=0pt, partopsep=0pt, itemsep=0pt, parsep=0pt]
    \item[\textsymking\ ] Model \( A \) beats model \( B \) and model \( B \) beats model \( C \) both with high win probabilities (\( P_{\text{win}} = 0.75 \)).
    \item[\textsymrook\ ] Model \( A \) beats model \( B \) with a high win probability (\( P_{\text{win}} = 0.75 \)), model \( B \) beats model \( C \) with a win probability close to 0.5 (\( P_{\text{win}} = 0.51 \)).
    \item[\textsymbishop\ ] Model \( A \) beats model \( B \) with a win probability close to 0.5 (\( P_{\text{win}} = 0.51 \)), model \( B \) beats model\( C \) with a high win probability (\( P_{\text{win}} = 0.75 \)).
    \item[\textsymknight\ ] Model \( A \) beats model \( B \) with a win probability of 0.54, model \( B \) beats model \( C \) with a win probability of 0.51. 
\end{enumerate}

In each of these scenarios, we simulate matches for paired comparisons "\(A \) vs. \(B \)" and "\(B \) vs. \(C \)" and then rearrange these matches in an arbitrary order to form our baseline sequence.
This approach mimics how Elo ratings are computed for online leaderboards in the evaluation of large language models~\citep{chatArena,lin2023llmeval}.
We then analyze whether Elo scores maintain the expected model hierarchies.

\begin{figure}[t]
  \centering
  \captionsetup{justification=centering}
  \begin{subfigure}[b]{.48\textwidth}
    \centering
    \includegraphics[width=\linewidth]{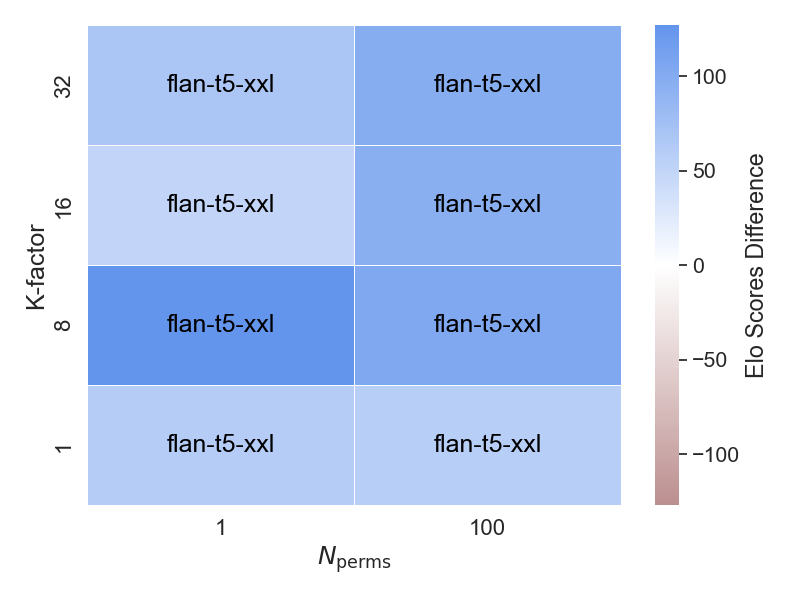}
    \caption{Experiment: Flan-t5-xxl vs. Flan-t5-xl \\ \textbf{Recorded Win rates}: \fbox{0.64 vs 0.36}}
    \label{fig:flan_heatmap}
  \end{subfigure}
  \hfill
  \begin{subfigure}[b]{.48\textwidth}
    \centering
    \includegraphics[width=\linewidth]{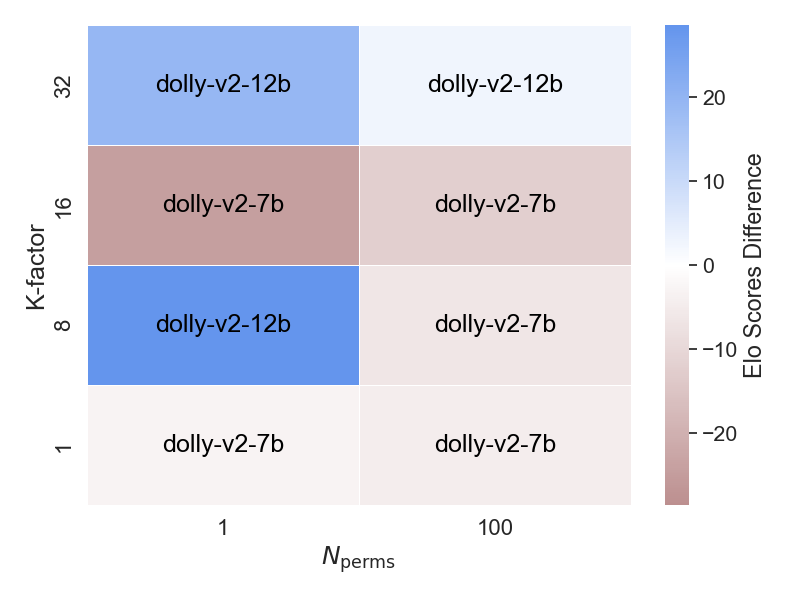}
    \caption{Experiment: Dolly-v2-7b vs. Dolly-v2-12b \\ \textbf{Recorded Win rates}: \fbox{0.51 vs 0.49}}
    \label{fig:dolly_heatmap}
  \end{subfigure}
  \caption{Final Elo scores difference (\(S_A - S_B\)) as a function of K-factor and \(N_{\text{perms}} \). This figure compares Model \( A \) (Flan-t5-xxl) and Model \( B \) (Flan-t5-xl). Positive values indicate the expected ranking where Model \( A \) is superior to Model \( B \), while negative values indicate a discrepancy, falsely suggesting that Model \( B \) has a higher Elo score than Model \( A \).}
\end{figure} 

\subsubsection{Key Findings}
The results of all 4 scenarios are consolidated in Table \ref{tab:elo-trans-synth}. 
These outcomes validate that the transitivity assumed by the Elo rating system can be vulnerable, especially when win rates hover around \( \approx 50\% \).
Once again, we observe that varying the number of permutations (\( n = 1 \) vs \( N_{\text{perms}} = 100 \)) and the \(K\)-factor plays a critical role in stability.
In the \textsymrook \space and \textsymbishop \space scenarios, with \( N_{\text{perms}} = 100 \) and \( K = 1 \), we notice discrepancies in the models' rankings.
This can be contrasted with \( K = 16 \), where rankings were much more consistent and reliable.
The slower updates from \( K = 1 \) suggest that this setting is possibly too conservative to capture the transitive relations quickly, hence leading to inconsistencies.


\begin{figure}[t]
  \centering
  \captionsetup{justification=centering}
  \begin{subfigure}[b]{.48\textwidth}
    \centering
    \includegraphics[width=\linewidth]{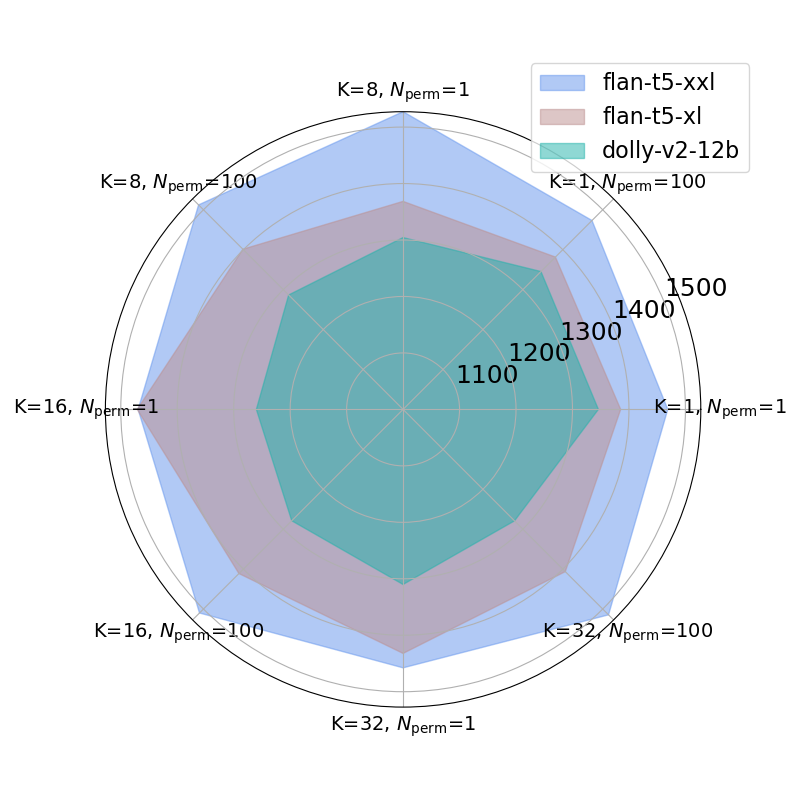}
    \caption{Flan-t5-xxl vs. Flan-t5-xl and Flan-t5-xxl vs. Dolly-v2-12b \\ \textbf{Recorded Win rates}: \fbox{0.64 vs 0.36} and \fbox{0.79 vs 0.21}}
    \label{fig:radar_flan_dolly}
  \end{subfigure}
  \hfill
  \begin{subfigure}[b]{.48\textwidth}
    \centering
    \includegraphics[width=\linewidth]{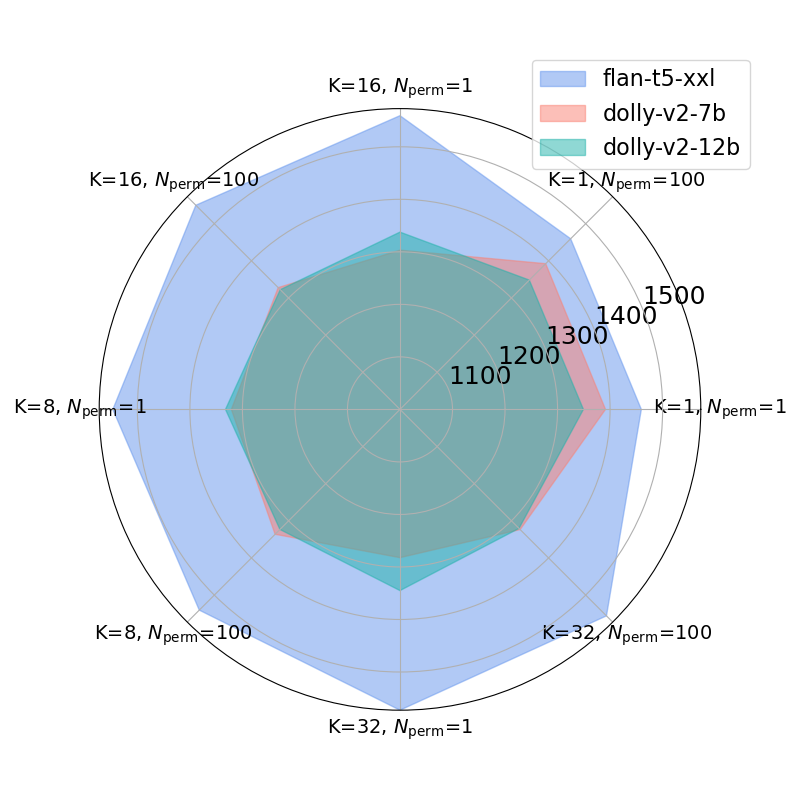}
    \caption{Dolly-v2-7b vs. Dolly-v2-12b and Flan-t5-xxl vs. Dolly-v2-12b  \\ \textbf{Recorded Win rates}: \fbox{0.51 vs 0.49} and \fbox{0.79 vs 0.21}}
    \label{fig:radar_dolly_flan}
  \end{subfigure}
  \caption{Final Elo scores (\(S_A\), \(S_B\) and \(S_C\)) for three different models at multiple configurations of \(N_{perms} = \{1, 100\}\) and \(K\)-factor = \(\{1, 8, 16, 32\}\). Intersecting of surfaces of individual model scores signifies that the relative ranking of the models is sensitive to these configurations. The order of model overlaps represent these models ranking based on their Elo scores.}
\end{figure}

\begin{table}[h!]
\centering
\caption{Win rates per evaluated model across conducted paired comparison experiments.}
\label{tab:win_rates}
\begin{tabular}{lc}
\hline
Experiment  & Win Rate\\
\hline
Flan-t5-xxl & 0.79 \\
Dolly-v2-12b & 0.21 \\
\hline
Flan-t5-xxl&  0.64 \\
Flan-t5-xl & 0.36 \\
\hline
Dolly-v2-7b & 0.51 \\
Dolly-v2-12b & 0.49 \\
\hline
\end{tabular}
\end{table}

\section{Validation on Real-World Human Feedback}
\label{sec:appl}

Building on the insights gained from our synthetic data experiments, we extend our validation efforts to include real-world human feedback.
Our objective is two-fold: first, to ascertain how the demonstrated properties established using synthetic data generalize to real human annotations; and second, to evaluate the Elo rating system's utility for assessing large language models (LLMs) under practical conditions.

\subsection{Experimental Setup}
Our study leverages human feedback data previously collected to explore data prioritization in language model evaluations. For details about our pool of prompts and models, completion generation, and annotation collection process, we refer the reader to the experimental setup section of our previous work~\citep{boubdir2023prompts}.
We focus on models from the well-established Dolly~\citep{DatabricksBlog2023DollyV2} and Flan~\citep{T5Chung2022} families, ensuring relevance to the broader NLP community.
The evaluation dataset consists of 400 prompts, with 100 randomly chosen from the \textsc{SODA} \citep{kim2022soda} dataset and 100 from each of the \textsc{CommonsenseQA}  \citep{talmor-etal-2019-commonsenseqa}, \textsc{CommonGen} \citep{lin-etal-2020-commongen}, and \textsc{AdversarialQA} \citep{Bartolo2020BeatTA} subsets, all of which are part of the Public Pool of Prompts (P3) dataset~\citep{sanh2021multitask}.
This ensures a diverse set of evaluation scenarios for a comprehensive assessment of the models’ capabilities.
Consistent with our synthetic data methodology, tie outcomes have been excluded from this analysis to focus specifically on the implications for the robustness of Elo scores.

In line with our previous analyses, we continue to explore the influence of variations in \(N_{\text{perms}} = \{1, 100\}\) and the \(K\)-factor (ranging from 1 to 36) on the robustness and reliability of Elo scores.
The win rates for each model, derived from human evaluations, are summarized in Table~\ref{tab:win_rates}.
Our real-world experiments yield two distinct types of scenarios: i) one in which a model decisively outperforms the other, such as the Flan-t5-xxl vs. Flan-t5-xl pairing; and ii) another one with two models nearly evenly matched, as in the Dolly-v2-7b vs. Dolly-v2-12b case.

\subsection{Key Findings}

Our analysis of real-world human feedback data reveals that the stability of Elo ratings is influenced by the disparities in win rates and the choice of hyperparameters \(K\)-factor and \(N_{\text{perms}}\).
In situations where win rates show a significant discrepancy, such as in our Flan family experiment, Elo ratings remain notably consistent across different \(K\)-factors and \(N_{\text{perms}}\) configurations (see Figure \ref{fig:flan_heatmap}).
On the other hand, in cases like the Dolly family experiment where win rates are closely matched, the Elo rating system exhibits higher volatility at \(N_{\text{perms}} = 1\) but gains stability at \(N_{\text{perms}} = 100\) at relatively small \(K\)-factors (see Figure \ref{fig:dolly_heatmap}).

Regarding the conservation of transitivity, our findings indicate that this property is not universally maintained in real-world human evaluations, as observed in synthetic data in Section \ref{sec:stress-tests}.
The relative rankings of models that perform similarly are sensitive to the choice of hyperparameters \(K\)-factor and \(N_{\text{perms}}\).
Consequently, one should exercise caution in drawing conclusions from the Elo scores when comprehensive paired comparison data, as dictated by the combination formula \ref{eq:combination}, is not available.
Our observations are in line with the trends seen in our synthetic data experiments.


\section{Empirical Guidelines for Robust Elo-based Evaluation}
\label{sec:guidelines}

In this section, we distill essential practices for enhancing the reliability of Elo-based evaluation of language models. 
These guidelines, derived from our empirical findings, differ notably from some conventional Elo settings and have significant implications for current real-world applications:

\begin{itemize}[partopsep=0pt]
    \item \textbf{Achieving Score Stability}: To obtain stable and reliable Elo ratings, it's recommended to run numerous permutations, ideally with \( N_{\text{perm}} \geq 100 \). This approach significantly improves the consistency of outcomes over single or fewer permutations commonly used.
    
    \item \textbf{Adjusting the \(K\)-factor}: A smaller K-factor may reduce significant rating fluctuations when models have closely matched win rates.
    
    \item \textbf{Rapid Convergence for Clear Winners}: When there's a clear performance disparity between models, a higher K-factor accelerates the alignment of Elo ratings with the models' ``true'' performance levels. This is in stark contrast to traditional uses of Elo ratings, where a one-size-fits-all K-factor is frequently applied.
    
    \item \textbf{Transitivity is not guaranteed}: The assumption that (\( A \text{ beats } B \) and \( B \text{ beats } C \) implies \( A > C \)) is not consistently valid in Elo ratings.This is particularly invalid when models have similar performance levels, challenging a common assumption in many Elo-based evaluations.
\end{itemize}

These guidelines serve as empirically-grounded recommendations to improve the robustness and interpretability of Elo-based evaluations for LLMs.
Following these best practices will help in yielding more reliable conclusions on models' performance via human judgment.

\section{Related Work}

Several works have proposed improvements to Elo scores. Variants such as Glicko \citep{glickman1995,glickman1999parameter,glickman2012example} and TrueSkill\textsuperscript{\texttrademark} \citep{NIPS2006_f44ee263, minka2018trueskill} have incorporated more complex statistical methods into the original Elo framework, to address some of the limitations of the Elo rating system, particularly in the context of games with more than two players or teams, or games with more complex outcomes than just win or loss.
There is also ongoing research into the efficacy of these systems in diverse and dynamic environments~\citep{Dehpanah2021EvaluatingTS,pmlr-v206-bertrand23a}. Prior work has demonstrated some limitations of Elo in maintaining transitivity, especially in non-transitive cyclic games such as rock-paper-scissors and StarCraft \textsc{ii} ~\citep{pmlr-v206-bertrand23a,vadori2023ordinal}. However, our work diverges by focusing on the reliability of Elo applied to large language model systems. To-date there has not been a comprehensive evaluation in this context. 

Independent from Elo, numerous studies have explored how sensitivity to hyperparameters can undermine the generalization of findings~\citep{novak2018sensitivity,lucic2018gans,Henderson2017,kadlec2017,MLSYS2021_cfecdb27} in machine learning. This forms part of a wider body of work which considers which factors influence reliability and reproducibility \citep{Goodman341ps12,Gundersen2018StateOT,barba2018terminologies,risnotr}. Notable directions includes studies on the impact of random seeds \citep{nagarajan2018impact,madhyastha-jain-2019-model,Summers2021NondeterminismAI}, model design choices \citep{shamir2020smooth,Snapp2021,pozzobon2023challenges,ko2023fairensemble}, the use of data parallelism \citep{shallue2019measuring}, hardware \citep{Zhuang2021RandomnessIN} and test set construction \citep{sogaard-etal-2021-need,lazaridou2021pitfalls,Melis2018OnTS}. Our work is complementary to these efforts, providing a rigorous evaluation of the impact of key hyperparameters and experimental settings on Elo performance.
\section{Conclusion}
This paper provides a comprehensive study on the reliability of the Elo rating system for evaluating LLMs using human feedback using an axiomatic framework. We identify various factors that influence the robustness of Elo ratings and offer guidelines for their effective application in real-world scenarios.
While our findings lay down an essential framework, they are by no means exhaustive. Future work could extend the present study by considering tie outcomes and adopting multi-category Bernoulli synthetic data to more closely simulate the varied landscape of human feedback. Such extensions could provide additional insights into the convergence properties of the Elo rating system in the fast-evolving landscape of language models.

\bibliography{anthology,main}

\newpage
\onecolumn
\appendix

\section{Extension to Multiple Outcomes}
For scenarios where outcomes can extend beyond wins and losses, such as a tie option, the multinomial distribution becomes relevant. For the outcomes win, loss, and tie, we sample according to the distribution:
\begin{align}
    P\left( n_{\text{win}}, n_{\text{loss}}, n_{\text{tie}} ; N, p_{\text{win}}, p_{\text{loss}}, p_{\text{tie}} \right) \notag \\
    = \frac{N!}{n_{\text{win}}! n_{\text{loss}}! n_{\text{tie}}!} p_{\text{win}}^{n_{\text{win}}} p_{\text{loss}}^{n_{\text{loss}}} p_{\text{tie}}^{n_{\text{tie}}}
\end{align}

\section{Impact of Ordering on Elo Ratings: Skewed Win Rates}
\label{apdx:elo-ordering-impact}
We summarize our findings on the impact of match sequences on Elo ratings for winning probabilities \(Prob(A \; \mathrm{{beats}} \; B) \geq 0.65\). 
\begin{figure*}[h!]
  \centering
    \includegraphics[width=1.\linewidth]{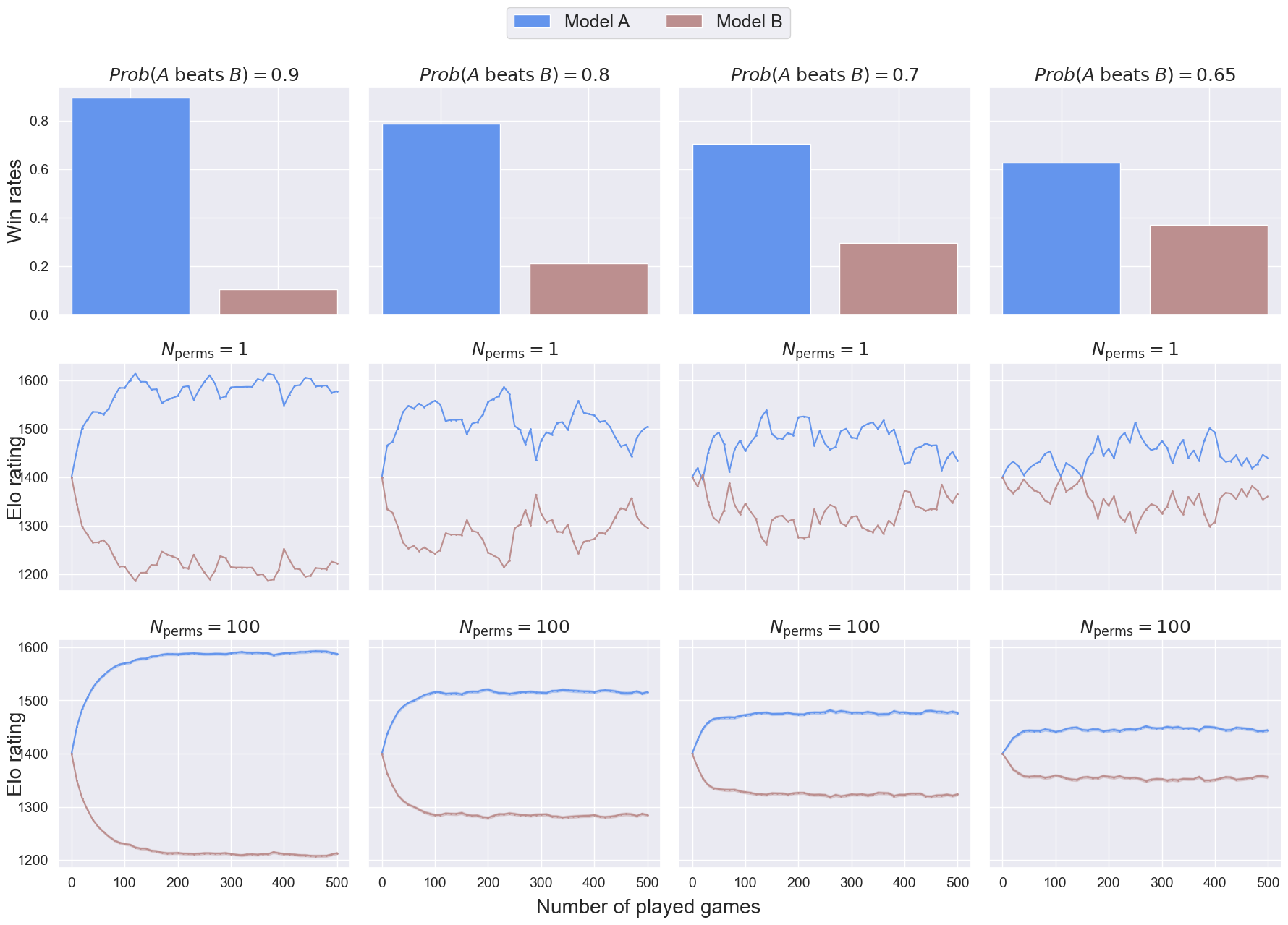}
\caption{\textbf{Impact of win probabilities and permutation sampling on Elo ratings}: Comparing Model A and Model B across three different win probabilities (\(Prob(A \; \mathrm{{beats}} \; B) = 0.9, 0.8, 0.7, 0.65\)) with two levels of permutation sampling (\(N_{\text{{perms}}}=1\) and \(N_{\text{{perms}}}=100\)). The top row displays the observed win rates, the middle row illustrates Elo ratings with a single permutation, and the bottom row shows the mean and standard error of the mean (SEM) of Elo ratings across 100 permutations.}
\label{fig:ordering-impact-2}
\end{figure*}

\newpage
\section{Experiment Flan-t5-xxl vs. Dolly-v2-12b Results}
\begin{figure*}[h]
  \centering
  \captionsetup{justification=centering}
  \includegraphics[width=0.65\linewidth]{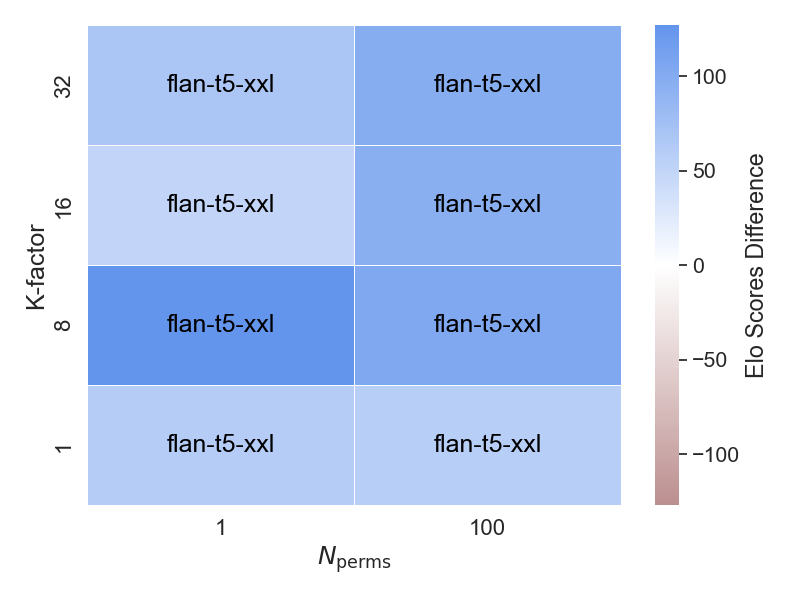}
  \caption{Experiment: Flan-t5-xxl vs. Dolly-v2-12b \\ \textbf{Recorded win rates}: \fbox{0.79 vs 0.21}}
  \label{fig:flan_heatmap}
\end{figure*}

\end{document}